\documentclass[letterpaper, 10 pt, conference]{ieeeconf}  

\IEEEoverridecommandlockouts                              

\usepackage{cite}
\usepackage{amssymb}
\usepackage{xcolor}
\usepackage{graphicx}
\usepackage{booktabs}
\usepackage{caption}
\usepackage{float}

\title{\LARGE \bf
Legs Over Arms: On the Predictive Value of Lower-Body Pose for Human Trajectory Prediction from Egocentric Robot Perception
}

\author{Nhat Le, Daeun Song,  and Xuesu Xiao
\thanks{All authors are with the Department of Computer Science, George Mason University } 
}

\begin{document}

\makeatletter
\g@addto@macro\@maketitle{
  \begin{figure}[H]
  \setlength{\linewidth}{\textwidth}
  \setlength{\hsize}{\textwidth}
  \centering
    \includegraphics[width=1\textwidth]{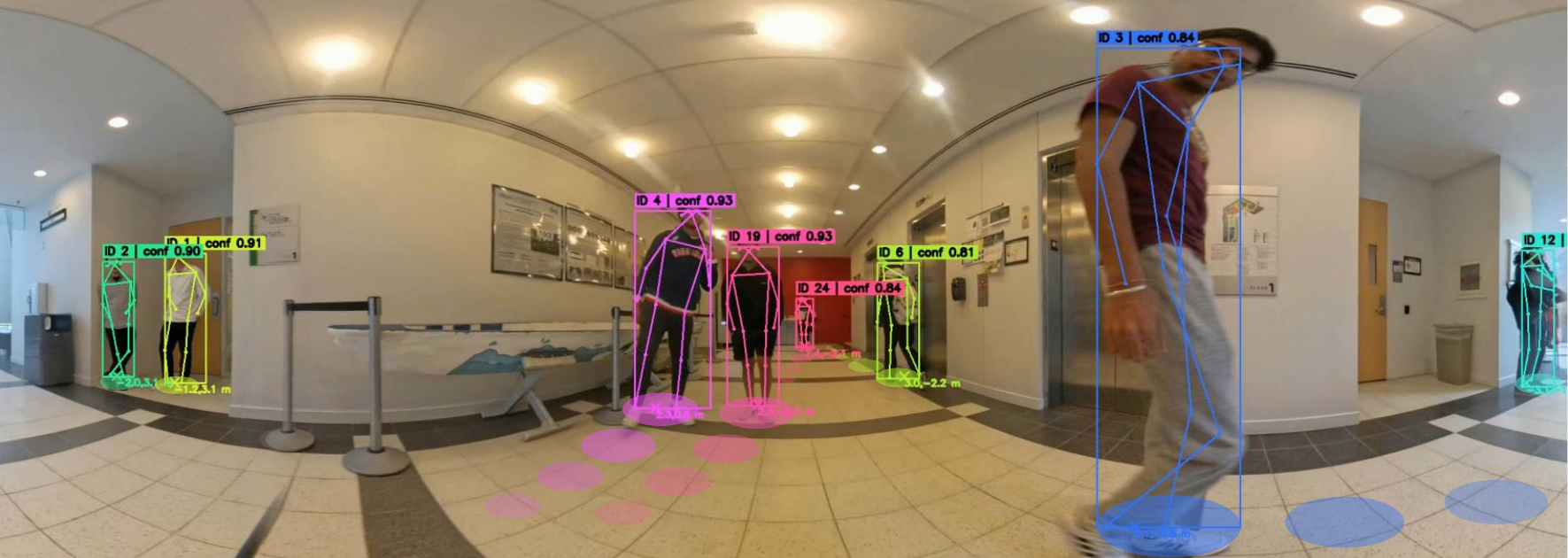}
    \caption{Human Trajectory Prediction Using Human Skeletal Keypoints from 360\textdegree~Robot Egocentric Camera View.}\vspace{-1.0em}
    \label{cover}
  \end{figure}
}
\makeatother
\maketitle
\addtocounter{figure}{-1}
\begin{abstract}
    Predicting human trajectory is crucial for social robot navigation in crowded environments. While most existing approaches treat human as point mass, we present a study on multi-agent trajectory prediction that leverages different human skeletal features for improved forecast accuracy. In particular, we systematically evaluate the predictive utility of 2D and 3D skeletal keypoints and derived biomechanical cues as additional inputs. Through a comprehensive study on the JRDB dataset and another new dataset for social navigation with 360\textdegree\ panoramic videos, we find that focusing on lower-body 3D keypoints yields a 13\% reduction in Average Displacement Error and augmenting 3D keypoint inputs with corresponding biomechanical cues provides a further 1-4\% improvement. Notably, the performance gain persists when using 2D keypoint inputs extracted from equirectangular panoramic images, indicating that monocular surround vision can capture informative cues for motion forecasting. Our finding that robots can forecast human movement efficiently by watching their legs provides actionable insights for designing sensing capabilities for social robot navigation.
\end{abstract}

\section{INTRODUCTION}

Accurate multi-agent trajectory prediction is a key enabler for social robot navigation. As robots are extensively deployed in human-centric environments such as hospitals, campuses, and public venues, it increases the demand for robots' ability to anticipate human trajectories. This capability is essential for navigating safely and efficiently while respecting social norms and comfort of nearby people~\cite{mavrogiannis_core_2023, francis2025principles, mirsky2024conflict, xiao2022motion}.

While much of the literature still represents humans as points on 2D maps or bird's-eye view (BEV) tracks, multi-agent trajectory prediction can benefit from human visual features. Data-driven methods~\cite{gupta2018social, xu2023social, yue2022human, salzmann2020trajectron++} perform well on BEV datasets but abstract away visual cues such as body pose, gaze, and gait. This abstraction is limiting for human-robot interaction with short histories and occlusions due to corners and doorways \cite{salzmann_robots_2023}. Recent work begins to close this gap by injecting human visual cues into forecasting models.

Similar to humans who use head and eye movement to observe the surrounding environments, robots in human-crowded environments can also benefit from surround-view perception to ensure safety and other social navigation objectives. While a single front-facing camera can miss people approaching from sides and behind with a limited field of view, a 360\textdegree\ RGB camera offers full coverage for perceiving nearby humans and capturing incoming trajectories from all directions. Recent development has made 360\textdegree\ RGB camera a low-cost sensor modality for mobile robots. Although its equirectangular images introduce distortion and horizontal wrap-around, these effects are well understood and tractable with standard engineering.

Motivated by the potential of rich first-person-view visual cues and 360\textdegree\ RGB camera (Fig.~\ref{cover}), this paper presents a study on which visual information, particularly skeletal regions and derived cues, matter the most for multi-agent trajectory prediction, and how useful 2D keypoints from equirectangular images are in practice. Based on Human Scene Transformer (HST) \cite{salzmann_robots_2023}, we compare the added values of skeletal keypoints and derived biomechanical gait cues from 2D and 3D pose for the trajectory prediction problem on two egocentric robot perception datasets.
From our study, we observe the following:
\begin{itemize}
    \item Lower-body 3D skeletal keypoints hold the most predictive values for our trajectory prediction task.
    \item 2D skeletal keypoints from equirectangular images improve trajectory prediction despite distortion.
\end{itemize}
We translate these findings into practical recommendations for feature selection and camera placement and outline future directions.

\section{Related work}

We review related work in human trajectory prediction in social navigation, human pose detection and intent prediction, and existing egocentric perception datasets. 

\subsection{Human Trajectory Prediction for Social Navigation}
Predicting human motion is crucial to social robot navigation. Mavrogiannis et al.~\cite{poddar_crowd_2023, stratton2026human} showed that predicting human trajectories can reduce discomfort around robots, although current models need significant improvements to increase robot navigation metrics. Deep learning approaches~\cite{gupta2018social, xu2023social, yue2022human, salzmann2020trajectron++} model multi-agent interactions effectively, but they represent humans as 2D points and are benchmarked on bird's-eye datasets like ETH/UCY \cite{5459260, lerner2007crowds} and SDD \cite{robicquet2016learning}. These approaches ignore rich human visual features from robot egocentric perspective which are potentially beneficial for predicting human motion. 

Salzmann et al. \cite{salzmann_robots_2023} introduced the Human Scene Transformer (HST), showing that full-body 3D keypoints and head orientation improve prediction from robot egocentric perception. Gao et al. \cite{gao2025social} proposed a portable pose encoder that boosts trajectory forecasting, but evaluated mainly on synthetic data or single-agent scenarios and focused on ego-motion of single agents. Taking full-body pose as input, however, no prior work has examined how different skeletal regions and derived biomechanical cues affect multi-agent human trajectory prediction in egocentric robot navigation. 

Based on these insights, our study is the first to systematically compare different skeletal regions and biomechanical cues for multi-agent trajectory prediction from robot egocentric perspective, aiming to identify which human body signals matter most for socially aware prediction.

\subsection{Human Pose Detection and Intent Prediction}
Human skeletal pose detection has become practical through 2D keypoint detectors~\cite{cao2019openpose, sun_deep_2019, xu2022vitpose}. These models localize skeletal keypoints from RGB image frames, which can then be lifted to 3D pose sequences by fitting parametric human body models \cite{grishchenko_blazepose_2022, loper2023smpl} or learning human motion representations \cite{zhu2023motionbert}. This process provides temporally consistent 3D skeletal keypoints usable for downstream tasks.

In autonomous driving, human pose is utilized to predict path direction and street-crossing intention. Minguez et al. \cite{quintero_minguez_pedestrian_2019} showed that focusing on leg and shoulder joints improves path direction prediction. Zhang et al. \cite{zhang_pedestrian_2022} and Li et al. \cite{li_interpretable_2024} demonstrated the usage of derived biomechanical cues for intention prediction. While informative, these studies focus on single-agent pedestrian direction and crossing intent in driving scenarios and do not address trajectory prediction.

\subsection{Egocentric Perception Datasets}

Several datasets such as JRDB \cite{martin-martin_jrdb_2023}, SCAND \cite{karnan2022socially}, MuSoHu~\cite{nguyen2023toward}, GND~\cite{liang2025gnd}, and CODa \cite{zhang2024toward} provide perception data from egocentric-view sensors. However, SCAND, CODa, most of MuSoHu, and part of GND do not provide full surround RGB views, which are crucial for detecting skeletal keypoints of all nearby humans. SCAND, MuSoHu, and GND additionally lack trajectory annotation and features relatively sparse interaction with humans. Although JRDB also lacks social navigation behaviors as the robot is often stationary, it remains a useful benchmark for our study with trajectory and keypoint annotations from full surround views. To better capture the dynamics of mobile service robots operating in human environments, we introduce in Section \ref{Datasets} a new dataset that emphasizes realistic navigation interactions with desired perception data.

\section{Methodology}
We formulate the problem of multi-agent trajectory prediction from egocentric perception and set the stage for our study on the values of different visual features in solving this prediction problem. We also present details of our datasets and implementations to facilitate our study. 
\begin{figure}[t]
    \centering
    \includegraphics[width=\linewidth]{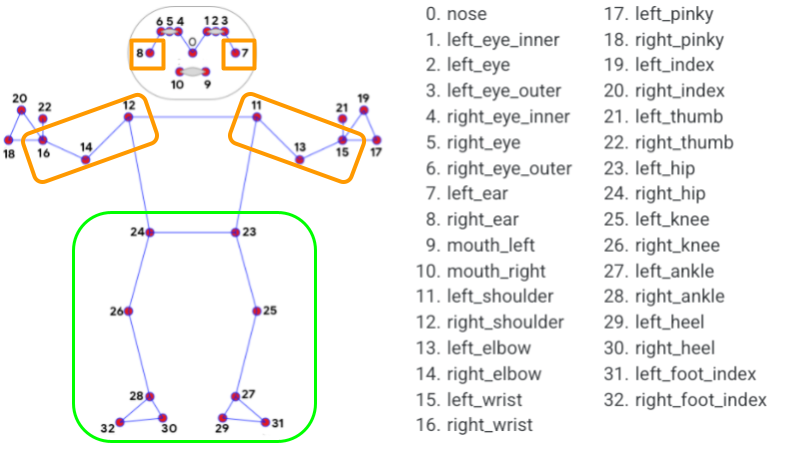}
    \caption{Diagram for 33-keypoint 3D skeletal pose with \textcolor{green}{$K^{3D}_L$} and \textcolor{orange}{$K^{3D}_U$} enclosed in green and orange boxes, respectively. Adapted from MediaPipePose\cite{mediapipe-pose}.}
    \label{fig3d}
\end{figure}

\subsection{Problem formulation}
For our study on the predictive value of skeleton-based keypoint features, we formulate the problem of multi-agent trajectory prediction from robot egocentric perception as follows: 
Let $\mathcal{A}_t$ denote the set of agents observed at time $t$. For each agent $i \in \mathcal{A}_t$, let $x_t^{(i)} \in \mathbb{R}^2$ be its 2D position in the robot-egocentric frame. Given a history window of length $H$, define the past trajectories
$$
X_{t-H+1:t} \;=\; \big\{\, x_{t-H+1:t}^{(i)} \,\big|\, i \in \mathcal{A}_t \big\}. \eqno{(1)}
$$
The prediction target is each agent's future positions over a horizon of $F$ steps,
$$
Y_{t+1:t+F} \;=\; \big\{\, x_{t+1:t+F}^{(i)} \,\big|\, i \in \mathcal{A}_t \big\}. \eqno{(2)}
$$
We learn a predictor $f_\theta$ that models a (potentially multimodal) distribution over futures conditioned on past motion and optional egocentric visual features:
$$
f_\theta\!\Big( Y_{t+1:t+F} \,\Big|\, X_{t-H+1:t},\, G_{t-H+1:t}(s) \Big), \eqno{(3)}
$$
where $G_{t-H+1:t}(s)$ denotes additional information detected from onboard perception (such as skeletal keypoints) and subsequently derived for observed agents based on feature configuration $s$. In particular, for observed agent $i \in \mathcal{A}_t$,
$$
G^{(i)}_{t-H+1:t}(s)=K^{(i)}_{t-H+1:t}(s) \cup C^{(i)}_{t-H+1:t}(s),
$$
where $K^{(i)}_{t-H+1:t}(s)$ denotes the set of keypoints available for configuration $s$ and $C^{(i)}_{t-H+1:t}(s)$ denotes corresponding derived cues. 

In our study with two datasets (see Section \ref{Datasets}), $G$ instantiates different feature configurations $s$ to enable controlled comparisons under a fixed training protocol. For 3D skeletal pose, we define $K^{3D}$ ,$K_L^{3D}$, $K_U^{3D}$ to be the set of all 33 keypoints, the subset of 10 lower-body keypoints, and the subset of 10 upper-body keypoints, respectively (see Fig. \ref{fig3d}). Subsequently, $C_L^{3D}$ derived from $K_L^{3D}$ includes leg articulation angles and step length, $K_U^{3D}$ includes arm articulation angles and head orientation derived from $K_U^{3D}$, and $C^{3D}=C_L^{3D}\cup C_U^{3D}$. For the 2D skeletal pose with standard COCO 17-keypoint format, we define $K^{2D}$ and $K_L^{2D}$ to be the set of all 17 keypoints and the subset of 6 lower-body keypoints from the hips down, respectively.

On JRDB, we conduct two sets of experiments. In the first experiment, we evaluate the impact of different 3D skeletal feature configurations to quantify the relative predictive value of skeletal regions and whether derived indicators provide complementary gains. In the second experiment, we directly compare 2D and 3D skeletal inputs, training on the publicly available 2D annotations from JRDB-Pose alongside 3D pose estimates, to assess how much predictive performance is lost when only 2D features are available.

For our new dataset, we focus on the predictive utility of 2D skeletal inputs extracted from equirectangular panoramic images. We specifically evaluate whether 2D keypoints from distorted panoramic imagery provide useful predictive cues, and whether restricting inputs to lower-body keypoints yields stronger improvements than using the full 2D skeleton. This tests our hypothesis that leg motion provides the most informative signals for short-term trajectory forecasting, even under uncorrected panoramic distortion.

\subsection{Datasets} \label{Datasets}

We base part of our evaluation on JRDB\cite{martin-martin_jrdb_2023}, a large-scale dataset collected from a mobile social robot with LiDAR and 360\textdegree\ stereo camera in pedestrian zones and indoor environments. JRDB provides multimodal egocentric perception and ground-truth annotations for human detection and tracking. A later version JRDB-Pose supplies 2D skeletal keypoints labels for all humans around the robot in COCO 17-keypoint format. Building on this foundation, HST used parametric human body fitting\cite{xu_ghum_2020, grishchenko_blazepose_2022} to estimate 3D skeletal keypoints for humans around the robots with a 33-keypoint format and made them publicly available. Although these methods may not be suitable for onboard real-time detection and tracking capabilities, they provide high-quality ground truth for skeletal representations that can convey intent and gait. We adopt JRDB and its pose extensions as a baseline to compare the gains from 2D and 3D skeletal features as well as derived mechanical cues and the predictive power of lower-, upper-, and full-body features for trajectory prediction. 

To complement JRDB, we additionally use a new social navigation dataset that will be released later. In this dataset, a teleoperated robot navigates through dense human crowds in indoor campus canteens and hallways in a socially compliant manner to reach goals implicitly determined by the operators. The platform is an AgileX Scout Mini equipped with an Insta360 X4 camera, producing equirectangular panoramic images for full surround-view coverage, along with other onboard sensors. We employ HRNet~\cite{sun_deep_2019} with MMPose~\cite{chen2019mmdetection} to detect and track humans and their 2D skeletal keypoints in image space, then convert the trajectories to robot Cartesian space using the pipeline by Bacchin et al.~\cite{bacchin2022people}, which leverages the properties of equirectangular images and the flat-ground assumption. This yields continuous 2D tracks of humans together with their 2D skeletal poses in COCO 17-keypoint format. Despite panoramic distortions near upper and lower image boundaries, the Insta360 X4 camera provides a full vertical field of view and reliably captures full-body poses at close range (see Fig. \ref{jrdb}). By contrast, the stitched 360\textdegree\ images in JRDB have a limited vertical field of view and often crop out the lower body and upper body of nearby pedestrians (see Fig. \ref{jrdb}).  In total, our dataset comprises 3 hours of socially-compliant and goal-oriented navigation, about three times the duration of JRDB. Unlike JRDB, where the robot frequently remains stationary or operates in relatively sparse outdoor settings, our dataset better reflects dynamic social navigation scenarios where the robot interacts with and navigates in dense human flows, making trajectory prediction a more challenging problem while imitating the real scenarios that mobile service robots encounter.

\subsection{Implementation Details}

\begin{figure}[t]
    \centering
    \includegraphics[width=0.9\linewidth]{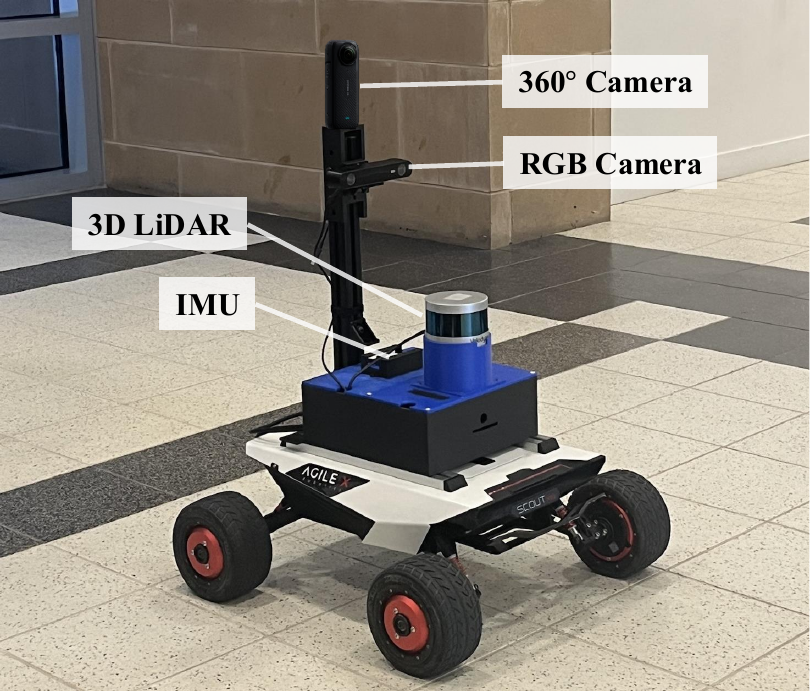}
    \caption{Our robot AgileX Scout Mini setup with onboard sensors: Insta360 X4 360\textdegree\ camera, Zed 2 RGB camera, Velodyne VLP-16 LiDAR, and WitMotion IMU.}
    \label{scout}
\end{figure}

\begin{figure*}[t]
    \centering
    \includegraphics[width=0.55\linewidth] {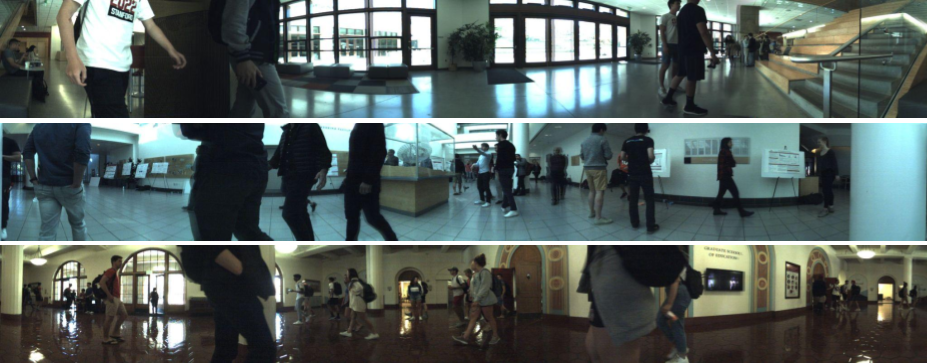} \includegraphics[width=0.43\linewidth]{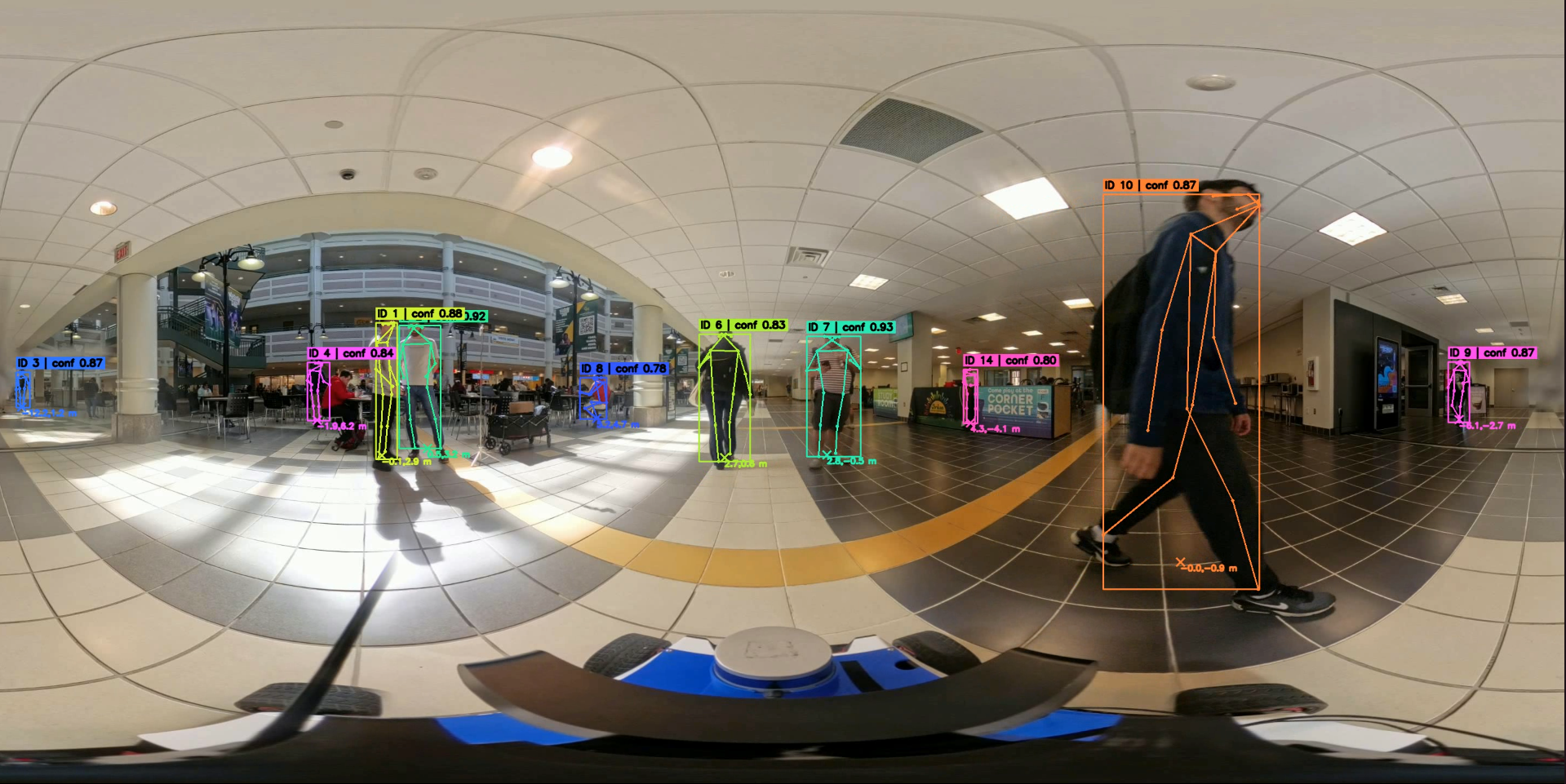}
    \caption{Sample panoramic images from JRDB (left) and our dataset (right). In JRDB, nearby humans are cropped due to the camera's limited vertical field of view, while in our dataset, full 2D keypoints can be detected even at distances below 1m.} \vspace{-0.5em}
    \label{jrdb}
\end{figure*}

We use the HST as the backbone for all experiments. It offers a flexible architecture for the multi-agent trajectory prediction problem as its modular design and attention mechanism make it straightforward to augment the model with additional modalities like skeletal keypoints and derived biomechanical cues. This extensibility allows us to perform controlled comparisons under a consistent modeling framework to ensure that observed differences in performance arise from the features themselves rather than architectural discrepancies. 

We re-implement the HST in PyTorch, closely following the design and reference code released with the original work. Each additional feature stream, such as skeletal keypoints, their subsets, and derived biomechanical cues, is encoded with a lightweight MLP before being fused with human trajectories for full self-attention across agent and timestep dimensions. Biomechanical cues include head orientation, step length, and limb articulation angles for both arms and legs, which serve as compact indicators of motion intent~\cite{prakash_recent_2018, hart2020using}. All models are trained with the AdamW optimizer, a batch size of 64, and configured to predict six consistent future modes per sequence.

For both datasets, we subsample the sequences to 3 Hz and train models to use information from six past timesteps (2s) to predict human trajectories in the next 12 timesteps (4s), ensuring consistent temporal resolution. We extract robot positions from odometry data and transform all trajectories into the robot’s egocentric frame at the first timestep of each sequence.

\subsection{Evaluation
}
During evaluation, we restrict metric computation to pedestrians for whom at least partial keypoint observations and position histories are available, ensuring feature parity across comparisons. We report our results using four trajectory prediction metrics, which capture geometric accuracy, quality of multimodal predictions, and probabilistic soundness of the predictive distribution:
\begin{itemize}
    \item Minimum Average Displacement Error ($MinADE$): the lowest average Euclidean distance between predicted and ground-truth trajectories across all sampled futures;
    \item Minimum Final Displacement Error ($MinFDE$): the lowest Euclidean distance between the predicted and ground-truth positions at the final timestep;
    \item Most Likely Average Displacement Error ($MLADE$): the average Euclidean distance between the most likely mode and ground-truth trajectories; and 
    \item Negative Log-Likelihood of Positions ($NLL_{pos}$): the negative log-likelihood of the ground-truth trajectory under the model’s predicted probabilistic distribution.
\end{itemize}

\section{Results and Discussion}

We present our experiment results and discuss the findings to make recommendations for future feature selection and camera placement for human trajectory prediction. 

\begin{table}[t]
\caption{Evaluation on 3D skeletal keypoints and cues on JRDB}
\label{table1}
\begin{center}
\begin{tabular}{ccccc}
\toprule
\multicolumn{1}{c}{$G(s)$} & $MinADE$ & $MinFDE$ & $MLADE$ & $NLL_{pos}$  \\ \midrule
$\varnothing$ (baseline)           & 0.39          & 0.64          & 0.71         & 0.86 \\
$K^{3D}$                 & 0.37          & 0.62          & 0.68        & 0.88 \\
$K^{3D}\cup C^{3D}$      & 0.36          & 0.60           & 0.62         & 0.80  \\
$K_U^{3D}$               & 0.38          & 0.58          & 0.83         & 0.82 \\
$K_U^{3D}\cup C_U^{3D}$  & 0.36          & 0.57          & 0.61         & 0.75 \\
$K_L^{3D}$               & \textbf{0.34} & \textbf{0.57} & \textbf{0.61}       & \textbf{0.65}\\
$K_L^{3D}\cup C_L^{3D}$  & \textbf{0.34}          & \textbf{0.57}          & 0.69         & 0.67 \\ \bottomrule
\end{tabular}
\end{center}
\end{table}

\begin{table}[ht]
\caption{Evaluation of 3D vs. 2D skeletal feature configurations on JRDB}
\label{table2}
\begin{center}
\begin{tabular}{ccccc}
\toprule
\multicolumn{1}{c}{$G(s)$} & $MinADE$ & $MinFDE$ & $MLADE$ & $NLL_{pos}$  \\ 
\midrule
$\varnothing$ (baseline)               & 0.42                           & 0.62                           & 0.72                          & 1.08                    \\
$K^{3D}$                  & 0.39                           & 0.56                           & 0.65                          & 0.88                    \\
$K_L^{3D}$                & \textbf{0.37}                           & \textbf{0.54}                           & \textbf{0.64}                          & \textbf{0.77}                    \\
$K^{2D}$                  & 0.41                           & 0.6                            & 0.74                          & 1.08                    \\
$K_L^{2D}$                & 0.41                           & 0.59                           & 0.72                          & 0.97                    \\ \bottomrule
\end{tabular}
\end{center}
\end{table}

\begin{table}[ht]
\caption{Evaluation of 2D keypoints detected from equirectangular images on our dataset}
\label{table3}
\begin{center}
\begin{tabular}{ccccc}
\toprule
\multicolumn{1}{c}{$G(s)$}    & $MinADE$ & $MinFDE$ & $MLADE$ & $NLL_{pos}$  \\
\midrule
$\varnothing$ (baseline)          & 1.1    & 1.44   & 1.86  & 3.1  \\
$K^{2D}$    & \textbf{1.02}   & \textbf{1.34}   & \textbf{1.6}   & \textbf{2.8}  \\
$K^{2D}_L$ & 1.03   & 1.35   & 1.73  & 2.86 \\ 
\bottomrule
\end{tabular}
\end{center}
\end{table}

\subsection{Predictive Value of Lower-body Keypoints} \label{predictive_value}

Table \ref{table1} presents the experimental results for different configurations of 3D skeletal keypoints on the JRDB dataset. The result shows that restricting additional inputs $G(s)$ to lower-body 3D keypoints $K^{3D}_L$ yields the largest combined gains over the baseline across all metrics. In particular, $K^{3D}_L$ reduces $MinADE$ by 13\%, $minFDE$ by 11\%, $MLADE$ by 14\%, and $NLL_{pos}$ by 24\%, indicating that leg motion features carry the strongest predictive signal for our trajectory prediction task. Full-body $K^{3D}$ and upper-body $K^{3D}_U$ configurations also outperform the baseline, but both trail $K^{3D}_L$. We further observe that augmenting keypoints with derived biomechanical cues improves the corresponding configurations by 1-4\%. This suggests that while the cues are beneficial, the primary information is already well captured by 3D keypoints encoding under this training regime.

\subsection{3D versus 2D Keypoints on JRDB} \label{2Dvs3D}
We compare the 3D and 2D skeletal inputs on a paired subset of JRDB in which both features are available to ensure parity. As shown in Table \ref{table2}, the 3D input provides better predictive value for the trajectory prediction task. Both 2D keypoints $K^{2D}$ and their lower subset $K^{2D}_L$ yield only marginal changes on $MinADE$, whereas $K^{3D}$ and $K^{3D}_L$ yields 7\% and 12\% respectively. This is consistent with what we observe in Section \ref{predictive_value}. Two factors likely explain the minimal 2D gains on JRDB: (i) training and evaluation are done on a smaller dataset, and (ii) JRDB's 360\textdegree\ camera often crops lower and upper body parts of nearby pedestrians, degrading the quality of image-based 2D pose more than that of 3D pose, which is obtained via sequence-level human parametric fitting.

\subsection{2D Keypoints from Equirectangular Images}
Table \ref{table3} reports results on our social navigation dataset collected with an Insta360 X4. Full-body 2D keypoints $K^{2D}$, extracted from equirectangular images without distortion correction, reduce $MinADE$ by 7\% relative to the baseline, with similar trends across other metrics.  We attribute part of these gains to full vertical field of view of equirectangular images, which captures complete pose even at close range, mitigating the issues observed on JRDB. However, the difference between $K^{2D}$ and $K^{2D}_L$ is not significant, which is consistent with what we see in Section \ref{2Dvs3D} and motivating future analysis on when using lower-body 2D keypoints is preferable.

\subsection{Discussion}

\subsubsection{Sensor placement for surround-view}
Fig.~\ref{scout} shows our robot hardware setup. The 360\textdegree\ camera is mounted at 0.85m from the ground, approximately half of average adult height. This placement reduces extreme equirectangular distortion near upper and lower image boundaries on nearby humans and likely contributes to the gains we see with 2D keypoints. Practitioners can take it into account together with this paper's key results to consider proper sensor placements that balance perception capabilities and human's comfort.

\subsubsection{Sensor choice} While a single off-the-shelf 360\textdegree\ camera is simple and low-cost, a multi-fisheye rig can integrate more easily with different robot form factors and retain maximum vertical field of view. For social navigation, this can improve leg visibility and tracking continuity, especially near corners and obstacles.

\subsubsection{Human 3D Keypoint Detection On Equirectangular Images}
Recent work~\cite{10.1145/3458709.3458986} outlines a practical path: learn an accurate fisheye-to-equirectangular projection model, fine-tune 2D keypoint detectors on synthetic panoramic data, and then perform 2D-to-3D lifting. This pipeline applies to both single 360\textdegree\ and multi-fisheye configurations and can improve 3D pose quality and achieve real-time processing capability for downstream tasks.

\subsubsection{Future Work}
While we standardize our evaluation on HST to isolate feature effects as embedded additional inputs are treated equally with simple encoders, our results could still be backbone-sensitive and confirming the results across distinct suitable backbones remains future work. Furthermore, our study does not explicitly consider other social cues such as gestures and group behaviors that could affect pose and alter feature importance. To explore further improvements,  one can consider using more complex encoding scheme for skeletal representations \cite{gao2025social, mohamed2020social, yan2018spatial}

\section{Conclusion}
This study investigates the predictive values of different sets of human skeletal keypoints for multi-agent trajectory prediction from robot egocentric perception. From the study, we observe two key findings. First, lower-body 3D skeletal keypoints provide the most predictive value. Incorporating the set of 3D lower-body keypoints yields the highest gain of 13\% $ADE$ reduction over a baseline that uses only 2D position histories for trajectory prediction task on JRDB. Using biomechanical cues such as limb articulation angles and head orientation, derived from corresponding sets of 3D skeletal keypoints, yields an additional improvement of 1-4\%. Second, 2D skeletal keypoints from equirectangular images from the 360\textdegree camera improve prediction, even with distortion. On our collected social navigation dataset, 2D body keypoints extracted from equirectangular panoramic images improve performance by 7\% over the same baseline. These findings indicate that skeletal keypoints, especially in lower-body region, and corresponding biomechanical cues capture important information about human motion for trajectory prediction. We expect this paper to serve as a reference for designing robust human trajectory prediction system for mobile robots to achieve more socially compliant navigation.

\addtolength{\textheight}{-1cm}

\bibliographystyle{IEEEtran}
\bibliography{references}

\end{document}